\documentclass{article}
\usepackage{spconf,amsmath,graphicx,hyperref}
\usepackage{cite}
\usepackage{hyperref}
\usepackage{amssymb}
\hypersetup{hidelinks}
\usepackage{graphicx}
\usepackage{xcolor}
\usepackage{etoolbox}

\title{Think Before You Restore: Risk-Aware Manchu Manuscript Restoration with Stroke-Guided Attention}
\name{
\begin{tabular}{c}
Mingqiu Liang$^{1,\dagger}$ \qquad
Dongdong Wang$^{2,\dagger}$ \qquad
Siyang Lu$^{1,*}$\\
Ting Huang$^{1}$ \qquad
Yingjun Qi$^{3}$
\end{tabular}
\thanks{
$^\dagger$Mingqiu Liang and Dongdong Wang contributed equally to this work.
$^*$Corresponding author: Siyang Lu (sylu@bjtu.edu.cn).
}
}

\address{
$^{1}$School of Computer Science and Technology, Beijing Jiaotong University,
Beijing, China\\
$^{2}$College of Design, Construction, and Planning, University of Florida,
Gainesville, FL, USA\\
$^{3}$School of Japanese Studies, Dalian University of Foreign Languages,
Dalian, China
}
\begin{document}
\ninept

\maketitle
\begin{abstract}
Full-page blind restoration of historical Manchu manuscripts is challenging due to scarce annotations, unknown degradation regions, and fragile connected strokes. Generic restoration models may improve visual quality but often modify intact content, leading to over-restoration. We propose \textbf{SAGE-Restore} (\textbf{S}troke-\textbf{A}ware \textbf{G}ated r\textbf{E}storation), a selective restoration framework that first assesses where restoration is needed and then uses this assessment to guide restoration candidate generation and pixel-level selection. Its encoder predicts patch-level repair probabilities from complementary appearance and stroke-structural cues to condition restoration candidate generation, while the corresponding repair logits are refined into a pixel-level soft gate that selectively controls where the restoration candidate is applied. We further introduce a fidelity-aware evaluation protocol that jointly measures degraded-region recovery, intact-content preservation, and their balance. SAGE-Restore achieves the highest R-Recovery (0.463) and RFS (0.626), while maintaining high U-Fidelity (0.968), demonstrating an effective balance between restoration and content preservation.








\end{abstract}
%


%






\section{Introduction}
\label{sec:introduction}

Historical document restoration aims to recover damaged textual structures while preserving original content and appearance. This is particularly challenging for historical Manchu manuscripts, where sparse and irregular degradations such as ink loss, corrosion, and local fractures disrupt fine, connected strokes and may resemble intact structures in local appearance, making stroke structure an important cue for deciding where restoration is needed. With degradation locations unknown in advance, restoration must often operate blindly on full manuscript pages without predefined damage masks~\cite{gsdm:24}.

\begin{table*}[t]
\centering
\caption{Comparison of full-reference (FR) and no-reference (NR)
image quality metrics.}
\label{tab:conventional_metric_observation}
\renewcommand{\arraystretch}{0.92}
\setlength{\tabcolsep}{1.4pt}
\small
\begin{tabular}{lcccccccccc|cccccc}
\hline
& \multicolumn{10}{c|}{Full-reference (FR)}
& \multicolumn{6}{c}{No-reference (NR)} \\
Method
& PSNR$\uparrow$
& SSIM$\uparrow$
& MAE$\downarrow$
& MSE$\downarrow$
& RMSE$\downarrow$
& MS-S$\uparrow$
& FSIM$\uparrow$
& GMSD$\downarrow$
& VIF$\uparrow$
& NCC$\uparrow$
& NIQE$\downarrow$
& BRIS$\downarrow$
& IL-N$\downarrow$
& MAN$\uparrow$
& MUSIQ$\uparrow$
& CLIP$\uparrow$ \\
\hline
Blind-Omni \cite{phutke2023blind}
& \textbf{23.68}
& \textbf{0.98}
& \textbf{0.01}
& \textbf{0.01}
& \textbf{0.07}
& \textbf{0.96}
& \textbf{0.95}
& \textbf{0.10}
& \textbf{0.89}
& \textbf{0.94}
& \textbf{3.93}
& 25.61
& 34.79
& 0.54
& 60.19
& 0.65 \\
DiffBIR \cite{diffbir:24}
& 17.26
& 0.58
& 0.06
& 0.02
& 0.14
& 0.81
& \textbf{0.89}
& 0.20
& 0.09
& 0.77
& 6.80
& \textbf{17.71}
& 40.01
& \textbf{0.59}
& \textbf{64.97}
& \textbf{0.77} \\
GSDM \cite{gsdm:24}
& 18.89
& 0.91
& 0.04
& 0.02
& 0.12
& 0.87
& 0.84
& 0.17
& 0.44
& 0.78
& 4.25
& 30.17
& \textbf{30.85}
& 0.47
& 61.08
& 0.68 \\
\hline
\end{tabular}
\end{table*}

Recent image restoration methods have shown strong performance using convolutional networks, vision Transformers, conditional prompts, and generative priors~\cite{restormer:22,nafnet:22,promptir:23,diffbir:24,hypir:25}. Document restoration methods further exploit stroke and structural information~\cite{textgestalt:22,docres:24,gsdm:24,hdr28k:25}, while gated and mask-aware architectures use spatial constraints to control image modification~\cite{gatedconv:19,mat:22}. However, blind full-page restoration introduces another challenge: most regions may already be intact and should not be changed. A model that aggressively repairs degraded content may therefore modify valid strokes and paper textures, leading to \emph{over-restoration}, whereas a conservative model may leave damaged strokes unrecovered, resulting in \emph{under-restoration}.


This problem is not well reflected by conventional image quality assessment (IQA). Full-reference metrics mainly measure global pixel or structural similarity~\cite{wang2004ssim,wang2003msssim,sheikh2006vif,zhang2011fsim,xue2014gmsd}, while no-reference metrics focus on perceptual quality~\cite{mittal2012brisque,mittal2013niqe,zhang2015ilniqe,ke2021musiq,yang2022maniqa,wang2023clipiqa}. As shown in our observations, favorable scores on these metrics do not necessarily indicate successful restoration: a method may obtain high similarity by making few changes, or achieve better perceptual quality while altering originally intact content. Document-specific metrics likewise focus primarily on overall quality, readability, or binarization accuracy~\cite{documentquality:18,shahkolaei2019document,ntirogiannis2013binarization}. These metrics do not explicitly distinguish recovery in degraded regions from unintended changes to intact regions.

We therefore formulate full-page blind Manchu manuscript restoration as a \emph{risk-aware selective restoration} problem, where successful restoration requires both recovering degraded structures and preserving intact content. To evaluate this trade-off, we introduce a fidelity-aware evaluation protocol that separately measures degraded-region recovery and intact-content preservation, together with their balance. Accordingly, we propose \textbf{SAGE-Restore} (\textbf{S}troke-\textbf{A}ware \textbf{G}ated r\textbf{E}storation), which uses stroke-guided repair assessment to determine where restoration is needed and pixel-level gating to selectively apply the generated restoration candidate.

The main contributions are summarized as follows:
\begin{itemize}
\item We characterize the recovery--preservation trade-off in historical Manchu manuscript restoration and introduce R-Recovery, U-Fidelity, and RFS to quantify degraded-region recovery, intact-content preservation, and their balance, respectively.

\item We propose \textbf{SAGE-Restore}, a selective restoration framework that couples stroke-guided repair assessment with pixel-level restoration selection to suppress over-restoration without requiring manually annotated masks during inference.

\item Extensive experiments demonstrate superior recovery--fidelity performance over existing restoration methods, while ablations verify the complementary contributions of Stroke-Guided Attention and pixel-level gating.
\end{itemize}

\section{Observations and Analysis}
\label{sec:observation}

\begin{figure}[h]
\centering
\includegraphics[width=0.47\textwidth]{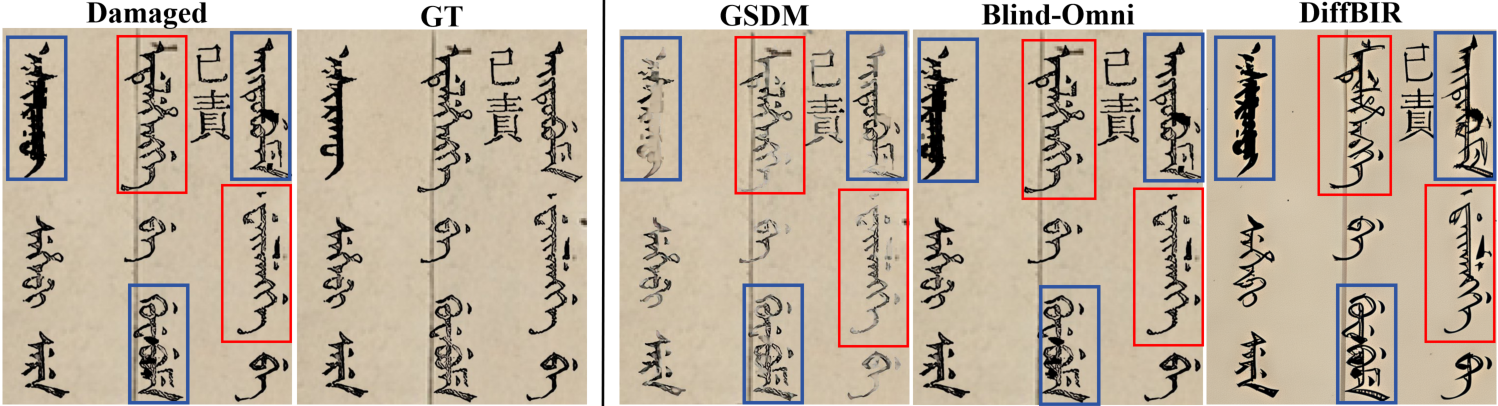}
\caption{Restoration performance across different methods, with red boxes indicating fidelity preservation and blue boxes indicating restoration quality.}
\label{fig:observation}
\end{figure}

We first evaluate the restoration quality of different methods using conventional image quality metrics. More importantly, we examine their fidelity in preserving intact content, which is critical for historical manuscript restoration. 
Table~\ref{tab:conventional_metric_observation} compares representative
full-reference (FR) and no-reference (NR) metrics~\cite{wang2004ssim,
wang2003msssim,zhang2011fsim,xue2014gmsd,sheikh2006vif,
mittal2012brisque,mittal2013niqe,zhang2015ilniqe,yang2022maniqa,
ke2021musiq,wang2023clipiqa} across restoration methods, while
Fig.~\ref{fig:observation} highlights the corresponding quality--fidelity
trade-off.

We first observe substantial differences in restoration quality across methods. DiffBIR is favored by several NR perceptual metrics, achieving the best MANIQA (MAN), MUSIQ, and CLIP scores, while Blind-Omni dominates the FR metrics. However, these metric scores do not fully reflect actual restoration behavior. As shown in Fig.~\ref{fig:observation}, Blind-Omni largely preserves the degraded input, leaving damaged strokes unrecovered, whereas DiffBIR provides stronger visual restoration. More importantly, stronger restoration can come at the cost of fidelity. GSDM and DiffBIR visibly modify originally intact strokes, introducing undesirable changes to valid content. This highlights a key limitation of conventional image quality metrics: they evaluate overall similarity or perceptual quality but do not explicitly assess the fidelity of intact regions.

These observations suggest that effective manuscript restoration should balance two complementary objectives: \emph{recovering degraded regions} while \emph{preserving intact regions}. Accordingly, we introduce \textbf{R-Recovery} to quantify restoration in degraded regions and \textbf{U-Fidelity} to measure fidelity in intact regions, and combine them into \textbf{RFS} for joint evaluation. Rather than favoring aggressive restoration or conservative preservation, RFS captures the desired principle of ``local restoration with global fidelity.''

Full-reference and no-reference metrics mainly assess global distortion or
perceptual quality, but cannot distinguish degraded-region recovery from
unintended changes to intact content. We therefore introduce three
region-aware metrics for evaluating restoration fidelity. Let $I_d$,
$\hat{I}$, and $I_{\mathrm{gt}}$ denote the degraded input, restored output,
and ground truth, respectively. Given the degradation mask $M_R$, the local
recovery rate is
$e_{\mathrm{in}}=\operatorname{MAE}(I_d,I_{\mathrm{gt}}\mid M_R)$,
$e_{\mathrm{out}}=\operatorname{MAE}(\hat{I},I_{\mathrm{gt}}\mid M_R)$, and
$R_{\mathrm{local}}=
\operatorname{clip}\!\left(
\frac{e_{\mathrm{in}}-e_{\mathrm{out}}}
{e_{\mathrm{in}}+\epsilon},0,1\right)$.
The intact region $M_U$ is divided into stroke and background regions,
$M_{U,S}$ and $M_{U,B}$. Its modification error and fidelity are
\begin{equation}
\begin{aligned}
E_U
&=\lambda\,\operatorname{MAE}(\hat{I},I_d\mid M_{U,S})\\
&\quad +(1-\lambda)\operatorname{MAE}(\hat{I},I_d\mid M_{U,B}),\\
U_F&=\operatorname{clip}(1-E_U,0,1),
\end{aligned}
\label{eq:u_fidelity}
\end{equation}
where $\lambda=0.8$ emphasizes intact-stroke preservation. R-Recovery and RFS
are defined as
\begin{equation}
R_R=R_{\mathrm{local}}U_F,\qquad
\mathrm{RFS}=\frac{2R_RU_F}{R_R+U_F+\epsilon},
\label{eq:fidelity_metrics}
\end{equation}
where $R_R$ and $U_F$ denote R-Recovery and U-Fidelity. Higher values indicate
a better balance between recovery and fidelity.


\begin{figure*}[t]
    \centering
    \includegraphics[width=\textwidth]{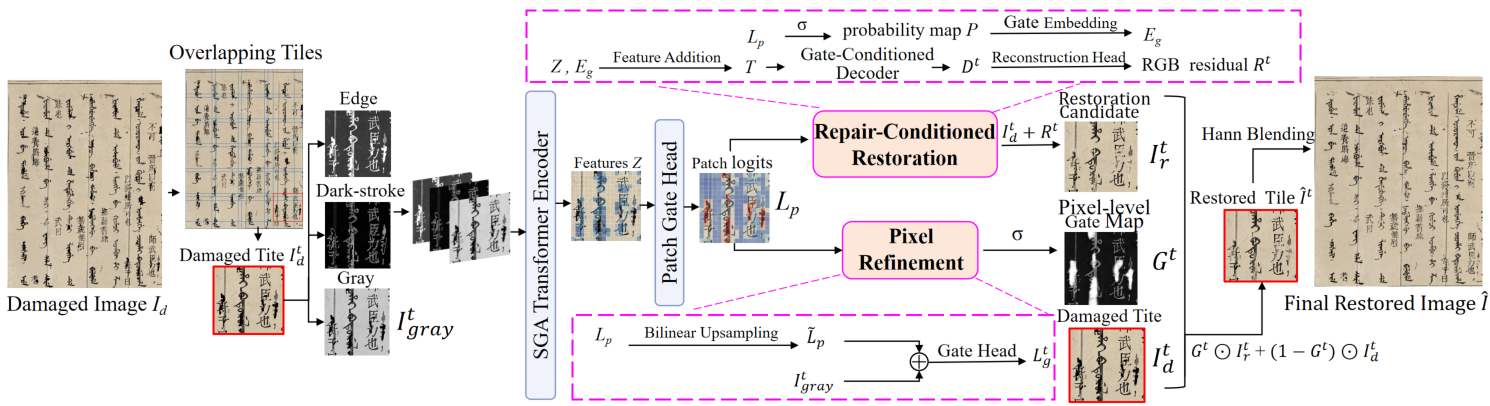}

    \caption{
Overview of the proposed SAGE-Restore framework.
The SGA encoder predicts patch-level repair logits \(L_p\), whose probabilities \(P\) condition restoration candidate generation.
The logits are further refined into a pixel-level soft gate \(G^t\), which selectively combines the degraded input \(I_d^t\) and restoration candidate \(I_r^t\).
Restored overlapping tiles are combined using Hann-weighted blending.
}
    \label{fig:framework}
\end{figure*}

\begin{figure*}[h]
\centering
\includegraphics[width=\textwidth]{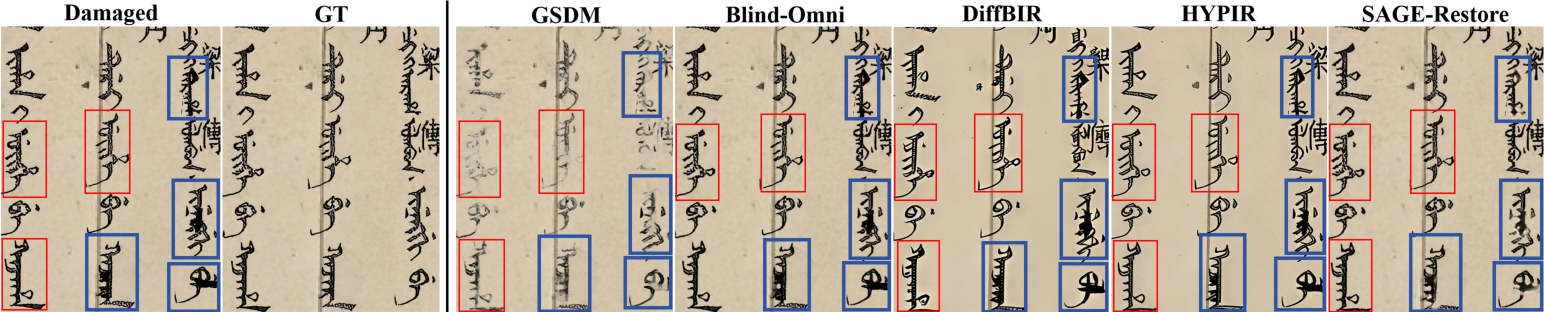}
\caption{
Qualitative comparison on two Manchu manuscripts.
GSDM recovers degraded strokes but introduces substantial changes to intact content, while Blind-Omni largely preserves the degraded input without recovering missing structures.
DiffBIR and HYPIR improve overall appearance but fail to reconstruct key missing strokes.
In contrast, SAGE-Restore selectively restores damaged structures while preserving intact handwriting and paper texture, achieving a better balance between recovery and fidelity.
}
\label{fig:qual-compare}
\end{figure*}









\begin{table*}[h]
\centering
\caption{Quantitative comparison with state-of-the-art restoration methods. Best results are in bold and second-best results are underlined.}
\label{tab:sota_comparison}
\renewcommand{\arraystretch}{0.95}
\setlength{\tabcolsep}{3.0pt}
\small
\begin{tabular}{lccc|cccccc|ccc}
\hline
& \multicolumn{3}{c|}{Fidelity-aware}
& \multicolumn{6}{c|}{Full-reference (FR)}
& \multicolumn{3}{c}{No-reference (NR)} \\
Method
& R-Rec.$\uparrow$
& U-Fid.$\uparrow$
& RFS$\uparrow$
& PSNR$\uparrow$
& SSIM$\uparrow$
& MAE$\downarrow$
& FSIM$\uparrow$
& GMSD$\downarrow$
& VIF$\uparrow$
& NIQE$\downarrow$
& MUSIQ$\uparrow$
& CLIP$\uparrow$ \\
\hline
Blind-Omni
& 0.033 & \textbf{0.993} & 0.063
& \underline{23.677} & \textbf{0.977} & \underline{0.010}
& \underline{0.954} & \underline{0.101} & \textbf{0.889}
& \textbf{3.932} & 60.188 & 0.653 \\
DiffBIR
& 0.062 & 0.848 & 0.116
& 17.264 & 0.576 & 0.057
& 0.886 & 0.204 & 0.088
& 6.795 & \textbf{64.967} & \underline{0.774} \\
GSDM
& \underline{0.420} & 0.819 & \underline{0.555}
& 18.888 & 0.913 & 0.040
& 0.837 & 0.171 & 0.442
& \underline{4.250} & \underline{61.080} & 0.683 \\
HYPIR
& 0.047 & 0.889 & 0.089
& 19.399 & 0.807 & 0.038
& 0.923 & 0.156 & 0.170
& 5.867 & 60.762 & \textbf{0.819} \\
SAGE-Restore
& \textbf{0.463} & \underline{0.968} & \textbf{0.626}
& \textbf{26.077} & \underline{0.975} & \textbf{0.006}
& \textbf{0.971} & \textbf{0.094} & \underline{0.856}
& 4.056 & 60.478 & 0.688 \\
\hline
\end{tabular}
\end{table*}

\section{Method}
\label{sec:method}
\subsection{Overall Framework}
\label{sec:framework}
We propose SAGE-Restore, a risk-aware selective restoration framework that couples stroke-guided repair assessment with restoration candidate generation and pixel-level selection. As shown in Fig.~\ref{fig:framework}, a shared repair assessment guides both where restoration corrections are generated and where they are applied, enabling damaged-region recovery while preserving intact content.
Given a degraded manuscript image $I_d$, we partition it into overlapping $512\times512$ tiles $\{I_d^t\}$ with a 64-pixel overlap. For each tile, grayscale, dark-stroke, and edge maps are divided into aligned $32\times32$ patches, forming a $16\times16$ token grid with 2-D positional encodings.
A Stroke-Guided Attention (SGA) encoder extracts stroke-aware features $Z$, from which a prediction head produces patch-level repair logits $L_p$ and probabilities
\begin{equation}
P=\sigma(L_p), \quad P\in[0,1]^{16\times16},
\label{eq:repair_prob}
\end{equation}
where $\sigma$ denotes the sigmoid function. The probabilities $P$ condition the restoration decoder to generate a candidate $I_r^t$, while the logits $L_p$ are refined into a pixel-level soft gate $G^t$. The final restored tile is
\begin{equation}
\hat I^t=(1-G^t)\odot I_d^t+G^t\odot I_r^t,
\label{eq:selective_restoration}
\end{equation}
where $\odot$ denotes element-wise multiplication. The gate selectively applies restoration corrections while retaining the original input in regions where restoration is unnecessary. Finally, restored tiles are combined using Hann-window weighted overlap blending to produce the full-page output $\hat I$.
\subsection{Stroke-Guided Repair Assessment}
\label{sec:repair_assessment}
Blind manuscript restoration requires distinguishing damaged regions from intact stroke structures with similar local appearances. We introduce SGA to incorporate stroke-related structural similarity and spatial proximity into patch interactions.

\noindent\textbf{Stroke-Aware Representation.} Grayscale, dark-stroke, and edge maps characterize local appearance, ink-like structures, and stroke boundaries, respectively. Their aligned patch embeddings provide complementary visual and structural cues.

\noindent\textbf{Stroke-Guided Attention.} SGA incorporates spatial and structural biases into content-based self-attention:
\begin{equation}
A_{\mathrm{SGA}}=\operatorname{Softmax}\left(\frac{QK^\top}{\sqrt d}+B_{\mathrm{sp}}+B_{\mathrm{str}}\right),
\label{eq:sga}
\end{equation}
where $Q$ and $K$ are query and key representations, and $d$ is their feature dimension. The biases are defined as
\begin{equation}
\begin{aligned}
s_i&=\operatorname{AvgPool}_i(0.7I_{\mathrm{dark}}+0.3I_{\mathrm{edge}}),\\
B_{\mathrm{str}}(i,j)&=1-|s_i-s_j|,\\
B_{\mathrm{sp}}(i,j)&=-\frac{\|p_i-p_j\|_1}{H_g+W_g-2},
\end{aligned}
\label{eq:sga_bias}
\end{equation}
where $s_i$ is the stroke-related descriptor of patch $i$, $p_i$ is its grid coordinate, and $H_g,W_g$ are the patch-grid dimensions. These biases favor interactions between structurally similar and spatially nearby patches, yielding stroke-aware encoder features $Z$.

\noindent\textbf{Patch-Level Repair Prediction.} A lightweight prediction head produces repair logits $L_p=H_p(Z)\in\mathbb{R}^{16\times16}$. The corresponding probabilities $P$ estimate patch-level restoration needs and condition candidate generation, while $L_p$ is retained for pixel-level gate prediction.
\subsection{Gate-Guided Restoration}
\label{sec:gate_restoration}
The repair assessment guides two complementary processes: restoration candidate generation and pixel-level selection.

\noindent\textbf{Restoration Candidate Generation.} The repair probabilities $P$ are projected into the token feature space through a gate-embedding MLP: $E_g=\operatorname{MLP}(P)$ and $T=Z+E_g$. Initialized with $T_0=T$, each decoder block applies self-attention, cross-attention with encoder features $Z$, and a feed-forward network. The resulting decoder features $Z_d$ are combined with the degraded tile to predict a bounded RGB residual:
\begin{equation}
\begin{aligned}
R_{\mathrm{raw}}^t&=H_{\mathrm{rec}}(Z_d,I_d^t),\\
\Delta^t&=r_{\max}\tanh(R_{\mathrm{raw}}^t),\\
R^t&=\begin{cases}
\Delta^t\odot(1-I_d^t), & \Delta^t\geq0,\\
\Delta^t\odot I_d^t, & \Delta^t<0.
\end{cases}
\end{aligned}
\label{eq:residual}
\end{equation}
The restoration candidate is obtained as $I_r^t=I_d^t+R^t$.

\noindent\textbf{Pixel-Level Gate Prediction.} The patch-level repair logits $L_p$ are bilinearly upsampled and refined using the grayscale degraded tile:
\begin{equation}
\begin{aligned}
\widetilde L_p&=\operatorname{Up}(L_p),\\
\Delta L_g^t&=H_{\mathrm{gate}}([\widetilde L_p,I_{\mathrm{gray}}^t]),\\
G^t&=\sigma(\widetilde L_p+\Delta L_g^t),
\end{aligned}
\label{eq:pixel_gate}
\end{equation}
where $G^t\in[0,1]^{H_t\times W_t\times1}$. The gate combines coarse repair assessment with high-resolution image information to selectively apply the restoration candidate according to Eq.~\eqref{eq:selective_restoration}.

\noindent\textbf{Training Objective.} We jointly optimize repair assessment and selective restoration using $\mathcal{L}=\mathcal{L}_{\mathrm{gate}}+\mathcal{L}_{\mathrm{recovery}}+\mathcal{L}_{\mathrm{preservation}}+\mathcal{L}_{\mathrm{aux}}$. The gate loss supervises patch-level repair prediction with focal BCE and
Tversky losses and pixel-level gate prediction with focal BCE. Recovery,
preservation, and auxiliary losses respectively apply structure-weighted
masked $\ell_1$ reconstruction, constrain both outputs to match the degraded
input in intact regions, and combine masked edge-gradient with global
$\ell_1$ losses.

\section{Experiments}
\label{sec:experiments}
\subsection{Experimental Setup}
\label{sec:experimental_setup}

\noindent\textbf{Dataset and Evaluation.} The dataset contains 600 training, 120 validation, and 120 test samples. The test set includes ink occlusions, text-like degradations, and clean hard negatives to assess recovery and preservation. At inference, only the degraded image $I_d$ is provided. All methods are evaluated on the same test set using the same fidelity-aware, full-reference, and no-reference metrics.

\noindent\textbf{Implementation Details.} We trained SAGE-Restore for 30 epochs using $512\times512$ crops and the AdamW optimizer. Positive-region crops were sampled with a probability of $0.8$. The batch size was 1, with four gradient-accumulation steps, and the encoder was frozen for the first three epochs. The initial learning rates were $10^{-5}$ for the encoder and $10^{-4}$ for the remaining network, with cosine annealing to a minimum of $10^{-6}$. The weight decay was $10^{-4}$, and the checkpoint with the lowest validation loss was selected for evaluation.

\subsection{Comparison with State-of-the-Art Methods}
\label{sec:sota_comparison}

\noindent\textbf{Recovery--Preservation Trade-off.} Table~\ref{tab:sota_comparison} reveals distinct restoration behaviors across competing methods. Blind-Omni achieves the highest U-Fidelity (0.993) but exhibits limited R-Recovery (0.033), indicating that preserving the degraded input alone is insufficient to recover missing stroke structures. In contrast, GSDM achieves substantially higher R-Recovery (0.420) at the cost of reduced U-Fidelity (0.819), illustrating the trade-off between damaged-region recovery and intact-content preservation. SAGE-Restore achieves the highest R-Recovery (0.463) while maintaining high U-Fidelity (0.968), resulting in the highest RFS (0.626). These results demonstrate that selective restoration can improve damaged-region recovery while limiting unintended modifications to intact content.

\noindent\textbf{Conventional Image Quality versus Restoration Effectiveness.} SAGE-Restore achieves the highest PSNR (26.080) and the lowest MAE (0.006), while Blind-Omni obtains a slightly higher SSIM (0.977 versus 0.975). However, the substantial difference in R-Recovery between these two methods (0.033 versus 0.463) shows that high global structural similarity does not necessarily indicate successful recovery of missing strokes. Similarly, DiffBIR and HYPIR exhibit low R-Recovery (0.062 and 0.047, respectively), despite producing restored outputs. These results highlight the need to evaluate damaged-region recovery and intact-content preservation explicitly, rather than relying solely on global image quality metrics.

\noindent\textbf{Qualitative Analysis.} As shown in Fig.~\ref{fig:qual-compare}, Blind-Omni largely preserves the degraded input but leaves missing strokes unrecovered. GSDM recovers damaged structures while introducing noticeable modifications to intact handwriting. DiffBIR and HYPIR improve overall visual appearance but fail to reconstruct some key missing strokes. In contrast, SAGE-Restore selectively restores damaged strokes while preserving surrounding handwriting and paper texture. These observations are consistent with the quantitative results and illustrate the importance of balancing localized restoration with intact-content preservation.

\subsection{Ablation Study}
\label{sec:ablation}
Table~\ref{tab:ablation} investigates the individual and complementary contributions of Stroke-Guided Attention (SGA) and pixel-level gating.

\noindent\textbf{Effect of Stroke-Guided Attention.} Without pixel-level gating, introducing SGA improves R-Recovery from 0.322 to 0.448 and U-Fidelity from 0.881 to 0.925. With gating enabled, SGA further improves R-Recovery from 0.405 to 0.463. These results indicate that stroke-guided structural context enhances damaged-region recovery while also contributing to intact-content preservation.

\noindent\textbf{Effect of Pixel-Level Gating.} Without SGA, introducing pixel-level gating increases U-Fidelity from 0.881 to 0.965 and R-Recovery from 0.322 to 0.405. When combined with SGA, gating further improves U-Fidelity from 0.925 to 0.968. These results indicate that pixel-level restoration selection helps suppress unintended modifications to intact regions while supporting damaged-region recovery.

\noindent\textbf{Complementary Contributions.} SAGE-Restore achieves the highest R-Recovery (0.463), U-Fidelity (0.968), and RFS (0.626) among all variants, together with the best PSNR, SSIM, and MAE. These improvements demonstrate the complementary roles of stroke-guided repair assessment and pixel-level gating in balancing damaged-region recovery and intact-content preservation.


\begin{table}[t]
\centering
\caption{Ablation results for SGA and pixel-level gating.}
\label{tab:ablation}
\setlength{\tabcolsep}{2.5pt}
\renewcommand{\arraystretch}{1.05}
\begin{tabular}{@{}cccccccc@{}}
\hline
SGA & Gate & R-Rec. & U-Fid. & RFS & PSNR & SSIM & MAE \\
\hline
-- & -- & 0.322 & 0.881 & 0.472 & 21.270 & 0.873 & 0.028 \\
\checkmark & -- & 0.448 & 0.925 & 0.596 & 22.980 & 0.914 & 0.014 \\
-- & \checkmark & 0.405 & 0.965 & 0.571 & 24.780 & 0.944 & 0.009 \\
\checkmark & \checkmark & \textbf{0.463} & \textbf{0.968} & \textbf{0.626} & \textbf{26.077} & \textbf{0.975} & \textbf{0.006} \\
\hline
\end{tabular}
\end{table}



\section{Conclusion}
We presented SAGE-Restore, a selective framework for blind historical Manchu manuscript restoration that combines stroke-guided repair assessment with gate-guided restoration. A fidelity-aware evaluation protocol was introduced to quantify damaged-region recovery, intact-content preservation, and their balance. Experimental results demonstrate improved recovery--preservation performance over competing methods, while ablation studies confirm the complementary contributions of SGA and pixel-level gating. Future work will address severe stroke loss and extend the framework to multilingual manuscripts.

\section{COMPLIANCE WITH ETHICAL STANDARDS}
This study involved no human participants or animals; therefore, ethical approval was not required.
\section{ACKNOWLEDGMENT}
\label{section:Acknowledgments}
This research was supported by the General Program of the National Natural Science Foundation of China (No.62376023).

\bibliographystyle{IEEEbib}
\bibliography{strings,refs}

\end{document}